\def\year{2017}\relax

\documentclass[letterpaper]{article}

\usepackage{aaai17}
\usepackage{times}
\usepackage{helvet}
\usepackage{courier}
\usepackage{url}            
\usepackage{nicefrac}       

\usepackage{amsmath}
\usepackage{amssymb}
\usepackage{graphicx}

\title{Learning Unitary Operators with Help From $\mathfrak{u}(n)$}

\begin{document}
%
\title{Learning Unitary Operators with Help From $\mathfrak{u}(n)$}
\author{
    Stephanie L. Hyland\textsuperscript{1, 2}, Gunnar R\"atsch\textsuperscript{1}\\
    \textsuperscript{1}Department of Computer Science, ETH Zurich, Switzerland\\
    \textsuperscript{2}Tri-Institutional Training Program in Computational Biology and Medicine, Weill Cornell Medical, New York\\
    \texttt{\{hyland, raetsch\}@inf.ethz.ch}
}

\maketitle
\begin{abstract}
A major challenge in the training of recurrent neural networks is the so-called vanishing or exploding gradient problem. The use of a norm-preserving transition operator can address this issue, but parametrization is challenging. In this work we focus on unitary operators and describe a parametrization using the Lie algebra $\mathfrak{u}(n)$ associated with the Lie group $U(n)$ of $n \times n$ unitary matrices. The exponential map provides a correspondence between these spaces, and allows us to define a unitary matrix using $n^2$ real coefficients relative to a basis of the Lie algebra. The parametrization is closed under additive updates of these coefficients, and thus provides a simple space in which to do gradient descent. We demonstrate the effectiveness of this parametrization on the problem of learning arbitrary unitary operators, comparing to several baselines and outperforming a recently-proposed lower-dimensional parametrization. We additionally use our parametrization to generalize a recently-proposed unitary recurrent neural network to arbitrary unitary matrices, using it to solve standard long-memory tasks.
\end{abstract}

\section{Introduction}
While recurrent neural networks (RNNs) are seeing widespread success across many tasks, the fundamental architecture presents challenges to typical training algorithms.
In particular, the problem of `vanishing/exploding gradients'~\cite{hochreiter1991untersuchungen} in gradient-based optimization persists, where gradients either vanish or diverge as one goes deeper into the network, resulting in slow training or numerical instability.
The long short-term memory (LSTM) network~\cite{hochreiter1997long} was designed to overcome this issue.
Recently, the use of norm-preserving operators in the transition matrix - the matrix of weights connecting subsequent \emph{internal} states - of the RNN have been
explored~\cite{arjovsky2015unitary,mikolov2014learning,le2015simple}. 
Using operators with bounded eigenvalue spectrum should, as demonstrated by~\citeauthor{arjovsky2015unitary}~\shortcite{arjovsky2015unitary}, bound the norms of the gradients in the network, assuming an appropriate non-linearity is applied.
Unitary matrices satisfy this requirement and are the focus of this work.

Imposing unitarity (or orthogonality) on this transition matrix is however challenging for gradient-based optimization methods, as additive updates typically do not preserve unitarity.
Solutions include \emph{re-unitarizing} after each batch, or using a \emph{parametrization} of unitary matrices closed under addition.
In this work we propose a solution in the second category, using results from the theory of Lie algebras and Lie groups to define a general parametrization of unitary matrices in terms of skew-Hermitian matrices (elements of the Lie algebra associated to the Lie group of unitary matrices). 
As explained in more detail below, elements of this Lie algebra can be identified with unitary matrices, while the algebra is closed under addition, forming a vector space over real numbers.

While we are motivated by the issues of RNNs, and we consider an application in RNNs, our primary focus here is on a core question: how can unitary matrices be learned?
\emph{Assuming} the choice of a unitary transition matrix is an appropriate modelling choice, the gradients on this operator should ultimately guide it towards unitarity, if it is possible under the parametrization or learning scheme used.
It is therefore useful to know which approach is best.
We distil the problem into its simplest form (a learning task described in detail later), so that our findings cannot be confounded by other factors specific to the RNN long-term memory task, before demonstrating our parametrization in that setting.

\subsection{Related work}
We draw most inspiration from the recent work of~\citeauthor{arjovsky2015unitary}~\shortcite{arjovsky2015unitary}, who proposed a specific parametrization of unitary matrices and demonstrated its utility in standard long-term memory tasks for RNNs. We describe their parametrization here, as we use it later. 
Citing the difficulty of obtaining a general and efficient parametrization of unitary matrices, they use the fact that the unitary group is closed under matrix multiplication to form a composite operator:
\begin{equation}
U = \mathbf{D}_3 \mathbf{R}_2 \mathcal{F}^{-1} \mathbf{D}_2 \Pi \mathbf{R}_1 \mathcal{F} \mathbf{D}_1
\label{eqn:arjovsky}
\end{equation}
where each component is unitary and easily parametrized:
\begin{itemize}
\item $\mathbf{D}$ is a diagonal matrix with entries of the form $e^{i \alpha}$, $\alpha \in \mathbb{R}$
\item $\mathbf{R}$ is a complex reflection operator; $\mathbf{R} = \mathbb{I} - 2\frac{\mathbf{v}\mathbf{v}^{\dagger}}{\| \mathbf{v} \|^2}$ ($\dagger$ denotes Hermitian conjugate)
\item $\mathcal{F}$ and $\mathcal{F}^{-1}$ are the Fourier and inverse Fourier transforms (or, in practice, their discrete matrix representations)
\item $\Pi$ is a fixed permutation matrix
\end{itemize}
In total, this parametrization has $7n$ real learnable parameters ($2n$ for each reflection and $n$ for each diagonal operator), so describes a subspace of unitary matrices (which have $n^2$ real parameters). 
Nonetheless, they find that an RNN using this operator as its transition matrix outperforms LSTMs on the adding and memory tasks described first in~\citeauthor{hochreiter1997long}~\shortcite{hochreiter1997long}.
This prompted us to consider other parametrizations of unitary matrices which might be more expressive or interpretable.

\citeauthor{mikolov2014learning}~\shortcite{mikolov2014learning} constrain a part of the transition matrix to be close to the identity, acting as a form of long-term memory store, while \citeauthor{le2015simple}~\shortcite{le2015simple} \emph{initialize} it to the identity, and then use ReLUs as non-linearities.
\citeauthor{henaff2016orthogonal}~\shortcite{henaff2016orthogonal} study analytic solutions to the long-term memory task, supporting observations and intuitions that orthogonal (or unitary) matrices would be appropriate as transition matrices for this task.
They also study initializations to orthogonal and identity matrices, and consider experiments where an additional term in the loss function encourages an orthogonal solution to the transition matrix, without using an explicit parametrization.
\citeauthor{saxe2013exact}~\shortcite{saxe2013exact} study exact solutions to learning dynamics in deep networks and find that orthogonal weight initializations at each layer lead to depth-independent learning (thus escaping the vanishing/exploding gradient problem).
Interestingly, they attribute this to the eigenvalue spectrum of orthogonal matrices lying on the unit circle.
They compare with weights initialized to random, scaled Gaussian values, which preserve norms in expectation (over values of the random matrix) and find orthogonal matrices superior.
It therefore appears that preserving norms is \emph{not} sufficient to stabilize gradients over network depth, but that the eigenvalue spectrum must also be strictly controlled.

In a related but separate vein, \citeauthor{krueger2015regularizing}~\shortcite{krueger2015regularizing} penalize the difference of difference of norms between subsequent hidden states in the network.
This is \emph{not} equivalent to imposing orthogonality of the \emph{transition} matrix, as the norm of the hidden state may be influenced by the inputs and non-linearities, and their method directly addresses this norm.

The theory of Lie groups and Lie algebras has seen most application in machine learning for its use in capturing notions of \emph{invariance}.
For example, \citeauthor{miao2007learning}~\shortcite{miao2007learning}, learn infinitesimal Lie group generators (elements of the Lie algebra) associated with affine transformations of images, corresponding to visual perceptual invariances. 
This is different to our setting as our generators are already known (we assume the Lie group $U(n)$) and wish to learn the coefficients of a given transformation relative to that basis set of generators. 
However, our approach could be extended to the case where the basis of $\mathfrak{u}(n)$ is \emph{unknown}, and must be learned. 
As we find later (appendix B), the choice of basis can impact performance, and so may be an important consideration. 
\citeauthor{cohen2014learning}~\shortcite{cohen2014learning} learn commutative subgroups of $SO(n)$ (known as toroidal subgroups), motivated by learning the irreducible representations of the symmetry group corresponding to invariant properties of images. Their choice of group parametrization is equivalent to selecting a particular basis of the corresponding Lie algebra, as they describe, but primarily exploit the algebra to understand properties of toroidal subgroups.

\citeauthor{tuzel2008learning}~\shortcite{tuzel2008learning} perform motion estimation by defining a regression function in terms of a function on the Lie algebra of affine transformations, and then learning this. 
This is similar to our approach in the sense that they do optimization in the Lie algebra, although as they consider two-dimensional affine transformations only, their parametrization of the Lie algebra is straight forward.

Finally, \citeauthor{warmuthorthogonal}~\shortcite{warmuthorthogonal} describe an online learning algorithm for orthogonal matrices -- which are the real-valued equivalent to unitary matrices. 
They also claim that the approach is extends easily to unitary matrices.

\subsection{Structure of this paper}
We begin with an introduction to the relevant facts and definitions from the theory of Lie groups and Lie algebras, to properly situate this work in its mathematical context.
Further exposition is beyond the scope of this paper, and we refer the interested reader to any of the comprehensive introductory texts on the matter.

We explain our parametrization in detail and describe a method to calculate the derivative of the matrix exponential - a quantity otherwise computationally intractable.
Then, we describe a simple but clear experiment designed to test our core question of learning unitary matrices. 
We compare to an approach using the parametrization of~\citeauthor{arjovsky2015unitary}~\shortcite{arjovsky2015unitary} and one using polar decomposition to `back-project' to the closest unitary matrix.
We use this experimental set-up to probe aspects of our model, studying the importance of the choice of basis (appendix B), and the impact of the restricted parameter set used by one of the alternate approaches.
We additionally implement our parametrization in a recurrent neural network as a `general unitary RNN', and evaluate its performance on standard long-memory tasks.

\section{The Lie algebra $\mathfrak{u}(n)$}
\label{section:lie}
\subsection{Basics of Lie groups and Lie algebras}
A Lie group is a group which is also a differentiable manifold, with elements of the group corresponding to points on the manifold. The group operations (multiplication and inversion) must be smooth maps (infinitely differentiable) back to the group. 
In this work we consider the group $U(n)$: the set of $n \times n$ unitary matrices, with matrix multiplication. These are the complex-valued analogue to orthogonal matrices, satisfying the property
\begin{equation}
U^{\dagger} U = U U^{\dagger} = \mathbb{I}
\label{eqn:unitary}
\end{equation}
where $\dagger$ denotes the conjugate transpose (or Hermitian conjugate). Unitary matrices preserve matrix norms, and have eigenvalues lying on the (complex) unit circle, which is the desired property of the transition matrix in a RNN.

The differentiable manifold property of Lie groups opens the door for the study of the Lie \emph{algebra}. This object is the \emph{tangent space} to the Lie group at the identity (the group must have an identity element).
Consider a curve through the Lie group $U(n)$ - a one-dimensional subspace parametrized by a variable $t$, where $U(t=0) = \mathbb{I}$ (this is a matrix $U(t)$ in $U(n)$ parametrised by $t$, not a group). Consider the defining property of unitary matrices (Equation~\ref{eqn:unitary}), and take the derivative along this curve:
\begin{equation}
U(t)^{\dagger}U(t) = \mathbb{I} \rightarrow \dot{U}(t)^{\dagger} U(t) + U^{\dagger}(t) \dot{U}(t) = 0
\end{equation}
Taking $t \rightarrow 0 $, $U(t) \rightarrow \mathbb{I}$, we have
\begin{equation}
\dot{U}(0)^{\dagger} \mathbb{I} + \mathbb{I}^{\dagger} \dot{U}(0) = 0 \Rightarrow \dot{U}(0)^{\dagger} = - \dot{U}(0)
\label{eqn:alg}
\end{equation}
The elements $\dot{U}(0)$ belong to the Lie algebra. We refer to this Lie algebra as $\mathfrak{u}(n)$, and an arbitrary element as $L$. Then Equation~\ref{eqn:alg} defines the properites of these Lie algebra elements; they are $n \times n$ skew-Hermitian matrices: $L^{\dagger} = -L$.

As vector spaces, Lie algebras are closed under addition. In particular $\mathfrak{u}(n)$ is a vector space over $\mathbb{R}$, so a \emph{real} linear combination of its elements is once again in $\mathfrak{u}(n)$ (this is also clear from the definition of skew-Hermitian). We exploit this fact later. 

Lie algebras are also endowed with an operation known as the \emph{Lie bracket}, which has many interesting properties, but is beyond the scope of this work. Lie algebras are interesting algebraic objects and have been studied deeply, but in this work we use $\mathfrak{u}(n)$ because of the \emph{exponential map}. 

Above, it was shown that elements of the algebra can be derived from the group (considering infinitesimal steps away from the identity). There is a \emph{reverse} operation, allowing elements of the group to be recovered from the algebra: this is the \emph{exponential map}. In the case of matrix groups, the exponential map is simply the \emph{matrix exponential}:
\begin{equation}
\exp(L) = \sum_{j = 0}^{\infty}\frac{L^j}{j!}
\end{equation}

Very simply, $L \in \mathfrak{u}(n)$, then $\exp(L) \in U(n)$. While this map is not in \emph{general} surjective, it so happens that $U(n)$ is a compact, connected group and so $\exp$ \emph{is} indeed surjective \cite{terrytao}. That is, for any $U \in U(n)$, there exists \emph{some} $L \in \mathfrak{u}(n)$ such that $\exp(L) = U$. Notably, while orthogonal matrices also form a Lie group $O(n)$, with associated Lie algebra $\mathfrak{o}(n)$ consisting of skew-symmetric matrices, $O(n)$ is \emph{not} connected, and so the exponential map can only produce \emph{special} orthogonal matrices - those with determinant one - $SO(n)$ being the component of $O(n)$ containing the identity.

\subsection{Parametrization of $U(n)$ in terms of $\mathfrak{u}(n)$}
The dimension of $\mathfrak{u}(n)$ as a real vector space is $n^2$. This is readily derived from noting that an arbitrary $n \times n$ complex matrix has $2n^2$ free real parameters, and the requirement of $L^{\dagger} = -L$ imposes $n^2$ constraints. So, a set of $n^2$ linearly-independent skew-Hermitian matrices defines a basis for the space; $\{ T_j \}_{j = \{1, \dots, n^2 \}}$. Then any element $L$ can be written as
\begin{equation}
L = \sum_{j=1}^{n^2} \lambda_j T_j
\end{equation}
where $\{\lambda_j\}_{j = 1, \dots, n^2}$ are $n^2$ real numbers; the coefficients of $L$ with respect to the basis. Using the exponential map,

\begin{equation}
U = \exp(L) = \exp\left(\sum_{j=1}^{n^2} \lambda_j T_j \right)
\label{eqn:parametrization}
\end{equation}
we see that these $\{\lambda_j\}_{j = 1, \dots, n^2}$ suffice as \emph{parameters} of $U$ (given the basis $T_j$). This is the parametrization we propose. It has two attractive properties:
\begin{enumerate}
\item It is a fully general parametrization, as the exponential map is surjective
\item Gradient updates on $\{\lambda_j\}_{j = 1, \dots, n^2}$ preserve unitarity automatically, as the algebra is closed under addition
\end{enumerate}
This parametrization means gradient steps are taken in the vector space of $\mathfrak{u}(n)$, rather than the manifold of $U(n)$, which \emph{may} provide a flatter cost landscape - although confirming this intuition would require further analysis. This work is intended to explore the use of this parametrization for learning arbitrary unitary matrices.

There are many possible choices of basis for $\mathfrak{u}(n)$. We went for the following set of sparse matrices:
\begin{enumerate}
\item $n$ diagonal, imaginary matrices: $T_a$ is $i$ on the $a$-th diagonal, else zero.
\item $\frac{n(n-1)}{2}$ symmetric, imaginary matrices with two non-zero elements, e.g., for $n=2$,
$\left(\begin{matrix} 0 &  i \\ i & 0 \end{matrix}\right)$
\item $\frac{n(n-1)}{2}$ anti-symmetric, real matrices with two non-zero elements, e.g., for $n=2$,
$\left(\begin{matrix} 0 &  1 \\ -1 & 0 \end{matrix}\right)$
\end{enumerate}

We explore the effects of choice of basis in appendix B.

\subsection{Derivatives of the matrix exponential}
The matrix exponential appearing in Equation~\ref{eqn:parametrization} poses an issue for gradient calculations. In general, the derivative of the matrix exponential does not have a closed-form expression, so computing gradients is intractable. 

In early stages of this work, we used the method of finite differences to approximate gradients, which would prohibit its use in larger-scale applications (such as RNNs). In the appendix we describe an investigation into using random projections to overcome this limitation, which while promising turned out to yield minimal benefit.

We therefore sought mathematical solutions to this complexity issue, which we describe here and in further detail in the appendix. Exploiting the fact that $L$ is skew-Hermitian, we can derive an analytical expression for the derivative of $U$ with respect to each of its parameters, negating the need for finite differences.

This expression takes the form:
\begin{equation}
\frac{\partial U}{\partial \lambda_a} = W V_a W^{\dagger}
\label{eqn:twomat}
\end{equation}
where $W$ is a unitary matrix of eigenvectors obtained in the eigenvalue decomposition of $U$; $U = W D W^{\dagger}$, ($D$ = diag($d_1$, $\dots$, $d_{n^2}$); $d_i$ are the eigenvalues of $U$).

Each $V_a$ is a matrix defined component-wise
\begin{align}
i = j:& V_{ii} =(W^{\dagger} T_a W)_{ii} e^{d_i}\label{eqn:vii}\\
i \neq j:& V_{ij} = (W^{\dagger} T_a W)_{ij} \left(\frac{e^{d_i} - e^{d_j}}{d_i - d_j}\right)\label{eqn:vij}
\end{align}
Where $T_a$ is the basis matrix of the Lie algebra in the $a$-th direction.

We provide the derivation, based on work from~\citeauthor{kalbfleisch1985analysis}~\shortcite{kalbfleisch1985analysis} and~\citeauthor{jennrich1976fitting}~\shortcite{jennrich1976fitting} in Appendix A.

We can simplify the expression $W^{\dagger} T_a W$ for each $T_a$, depending on the type of basis element. In these expressions, $\mathbf{w}_a$ refers to the $a$-th \emph{row} of W.
\begin{enumerate} 
\item $T_a$ purely imaginary; $W^{\dagger} T_a W = i \cdot \mathrm{outer}(\mathbf{w}^*_a, \mathbf{w}_a)$
\item $T_a$ symmetric imaginary, nonzero in positions $(r, s)$ and $(s, r)$: $W^{\dagger} T_{rs} W = i \cdot (\mathrm{outer}(\mathbf{w}^*_s, \mathbf{w}_r) + \mathrm{outer}(\mathbf{w}^*_r, \mathbf{w}_s))$
\item $T_a$ antisymmetric real, nonzero in positions $(r, s)$ and $(s, r)$: $W^{\dagger} T_{rs} W = \mathrm{outer}(\mathbf{w}^*_r, \mathbf{w}_s) - \mathrm{outer}(\mathbf{w}^*_s, \mathbf{w}_r)$
\end{enumerate}
These expressions follow from the sparsity of the basis and are derived in appendix A. Thus, we reduce the calculation of $W^{\dagger} T_a W$ from two matrix multplications to at most two vector outer products.

Overall, we have reduced the cost of calculating gradients to a single eigenvalue decomposition, and for each parameter two matrix multiplications (equation~\ref{eqn:twomat}), one or two vector outer products, and element-wise multiplication of two matrices (equations~\ref{eqn:vii},~\ref{eqn:vij}). As we see in the RNN experiments, this actually makes our approach faster than the (restricted)uRNN of~\cite{arjovsky2015unitary} for roughly equivalent numbers of parameters.

\section{Supervised Learning of Unitary Operators}
\label{section:learning}
We consider the supervised learning problem of
learning the unitary matrix $U$ that generated a $\mathbf{y}$ from
$\mathbf{x}$; $\mathbf{y} = U\mathbf{x}$, given examples of such
$\mathbf{x}$s and $\mathbf{y}$s. 
This is the core learning problem that needs to be solved for the state-transformation matrix in RNNs.
It is similar to the setting considered in \citeauthor{warmuthorthogonal}~\shortcite{warmuthorthogonal} (they consider an online learning problem). 
We compare a number of methods for learning $U$ at different values of $n$.
We further consider the case where we have artificially restricted the number of learnable variables in our parametrization (for the sake of comparison), and generate a pathological change of basis to demonstrate the relevance of selecting a good basis (appendix B).

\subsection{Task}
The experimental setup is as follows: we create a $n \times n$ unitary matrix $U$ (the next section describes how this is done), then sample vectors $\mathbf{x} \in \mathbb{C}^n$ with normally-distributed coefficients.
We create $\mathbf{y}_j = U \mathbf{x}_j + \epsilon_j$ where $\epsilon \sim \mathcal{N}(0, \sigma^2)$.
The objective is to recover $U$ from the $\{\mathbf{x}_j, \mathbf{y}_j\}$ pairs by minimizing the squared Euclidean distance between predicted and true $\mathbf{y}$ values;
\begin{equation}
U = \mathop{\mathrm{argmin}}_U \frac{1}{N} \sum_j^N \| \hat{\mathbf{y_j}} - \mathbf{y_j} \|^2 = \mathop{\mathrm{argmin}}_U \frac{1}{N} \sum_j^N \| U \mathbf{x}_j - \mathbf{y_j} \|^2
\end{equation}
While this problem is easily solved in the batch setting using least-squares, we wish to learn $U$ through mini-batch stochastic gradient descent, to emulate a deep learning scenario.

For each experimental run (a single $U$), we generate one million training $\{\mathbf{x}_j, \mathbf{y}_j\}$ pairs, divided into batches of size 20. The test and validation sets both contain $100,000$ examples. In practice we set $\sigma^2 = 0.01$ and use a fixed learning rate of $0.001$. For larger dimensions, we run the model through the data for multiple epochs, shuffling and re-batching each time.

All experiments were implemented in Python. The code is available here: \texttt{https://github.com/ratschlab/uRNN}. For the matrix exponential, we use the scipy builtin \texttt{expm}, which uses Pade approximation~\cite{al2009new}. We make use of the fact that $iL$ is Hermitian to use \texttt{eigh} (also in scipy) to perform eigenvalue decompositions.

\subsection{Generating the ground-truth unitary matrix}
\label{section:generating}

The $U$ we wish to recover is generated by one of three methods:
\begin{enumerate}
\item QR decomposition: we create a $n \times n$ complex matrix with normally-distributed entries and then perform a QR decomposition, producing a unitary matrix $U$ and an upper triangular matrix (which is discarded). This approach is also used to sample orthogonal matrices in~\citeauthor{warmuthorthogonal}~\shortcite{warmuthorthogonal}, noting a result from~\citeauthor{stewart1980efficient}~\shortcite{stewart1980efficient} demonstrating that this is equivalent to sampling from the appropriate Haar measure.
\item Lie algebra: given the standard basis of $\mathfrak{u}(n)$, we sample $n^2$ normally-distributed real $\lambda_j$ to produce $U = \exp{\left( \sum_j \lambda_j T_j\right)}$
\item Unitary composition: we compose parametrized unitary operators as in ~\citeauthor{arjovsky2015unitary}~\shortcite{arjovsky2015unitary} (Equation~\ref{eqn:arjovsky}). The parameters are sampled as follows: angles in $D$ come from $\mathcal{U}(-\pi, \pi)$. The complex reflection vectors in $\mathbf{R}$ come from $\mathcal{U}(-s, s)$ where $s = \sqrt{\frac{6}{2n}}$.
\end{enumerate}
We study the effects of this generation method on test-set loss in a later section. While we find no significant association between generation method and learning approach, in our experiments we nonetheless average over an equal number of experiments using each method, to compensate for possible unseen bias.

\subsection{Approaches}

We compare the following approaches for learning $U$:
\begin{enumerate}
\item \texttt{projection}: $U$ is represented as an unconstrained $n \times n$ complex matrix, but after each gradient update we \emph{project} it to the closest unitary matrix, using polar decomposition~\cite{keller1975closest}. This amounts to $2n^2$ real parameters.
\item \texttt{arjovsky}: $U$ is parametrized as in Equation~\ref{eqn:arjovsky}, which comes to $7n$ real parameters.
\item \texttt{lie\_algebra}: (we refer to this as $\mathfrak{u}(n)$) $U$ is parametrized by its $n^2$ real coefficients $\{ \lambda_j \}$ in the Lie algebra, as in Equation \ref{eqn:parametrization}.
\end{enumerate}

As baselines we use the \texttt{true} matrix $U$, and a \texttt{random} unitary matrix $U_R$ generated by the same method as $U$ (in that experimental run). 

We also implemented the algorithm described in~\citeauthor{warmuthorthogonal}~\shortcite{warmuthorthogonal} and considered both unitary and orthogonal learning tasks (our parametrization contains orthogonal matrices as a special case) but found it too numerically unstable and therefore excluded it from our analyses.

\subsection{Comparison of Approaches}
\label{section:results}

\begin{table*}
\small
\begin{tabular}{ r || c | c c  c | c}
$n$ & \texttt{true} & \texttt{projection}   & \texttt{arjovsky}                          & \texttt{lie algebra}                  &  \texttt{rand} \\ \hline
3 & $6.004 \pm 0.005 \times 10^{-4}$  & 8 $\pm$ 1             & $\mathbf{6.005 \pm 0.003 \times 10^{-4}}$  & $\mathbf{6.003 \pm 0.003 \times 10^{-4}}$    & $12.5 \pm  0.4$ \\
6 & $\sim$ 0.001    & $15 \pm 1$            & $0.09 \pm 0.01 $             & $\mathbf{0.03 \pm 0.01}$     & 24 $\pm$ 1 \\
8 & $\sim$ 0.002    & $14 \pm 1$            & $1.17 \pm 0.06$                               & $\mathbf{0.014 \pm 0.006}$  & $31.6 \pm 0.6$ \\
14 & $\sim$ 0.003   & $24 \pm 4$            & $10.8 \pm 0.3$                                & $\mathbf{0.07 \pm  0.02}$    & $52 \pm 1$ \\
20 & $\sim$ 0.004   & $38 \pm 3$            & $29.0 \pm 0.5$                                & $\mathbf{0.47 \pm 0.03}$     & $81 \pm 2$
\end{tabular}
\caption{Loss (mean $l_2$-norm between $\hat{y}_i$ and $y_i$) on the test set for the different approaches as the dimension of the unitary matrix changes. \texttt{true} refers to the matrix used to generate the data, \texttt{projection} is the approach of `re-unitarizing' using a polar decomposition after gradient updates, \texttt{arjovsky} is the composition approach defined in Equation~\ref{eqn:arjovsky}, $\mathfrak{u}(n)$ is our parametrization (Equation~\ref{eqn:parametrization}) and \texttt{rand} is a random unitary matrix generated in the same manner as \texttt{true}. Values in bold are the best for that $n$ (excluding \texttt{true}). The error for \texttt{true} is typically very small, so we omit it.}
\label{table:accuracies}
\end{table*}

Table~\ref{table:accuracies} shows the test-set loss for different values of $n$ and different approaches for learning $U$.
We performed between 6 and 18 replicates of each experiment, and show bootstrap estimates of means and standard errors over these replicates.
As we can see, the learning task becomes more challenging as $n$ increases, but our parametrization ($\mathfrak{u}(n)$) consistently outperforms the other approaches.

\subsection{Restricting to $7n$ parameters}
\label{section:restriction}
As mentioned, \texttt{arjovsky} uses only $7n$ parameters.
To check if this difference accounts for the differences in loss observed in Table~\ref{table:accuracies}, we ran experiments where we fixed all but $7n$ (selected randomly) of the $\{\lambda_j\}$ in the \texttt{lie\_algebra} parametrization. 
The fixed parameters retained their initial values throughout the experiment. We observe that, as suspected, restricting to $7n$ parameters results in a performance degradation equivalent to that of \texttt{arjovsky}.

\begin{table*}
\small
\begin{tabular}{r || c c | c}
$n$ & \texttt{arjovsky} & \texttt{lie\_restricted} & \texttt{lie\_unrestricted} \\\hline
8 & $1.2 \pm 0.1$ & $1.0 \pm 0.2$ & $0.04 \pm 0.01$\\
14 & $11.6 \pm 0.3$ & $ 12.6 \pm 0.4 $ & $0.25 \pm 0.03$ \\
20 & $27.8 \pm 0.7$ & $28.0 \pm 0.6$ & $0.19 \pm 0.03$
\end{tabular}
\caption{We observe that restricting our approach to the same number of learnable parameters as that of \cite{arjovsky2015unitary} causes a similar degradation in performance on the task. This indicates that the relatively superior performance of our model is explained by its generality in capturing arbitrary unitary matrices.}
\label{table:restrict}
\end{table*}

Table~\ref{table:restrict} shows the results for $n=8, 14, 20$.  The fact that the restricted case is consistently within error of the \texttt{arjovsky} model supports our hypothesis that the difference in learnable parameters accounts for the difference in performance. This suggests that generalising the model of \citeauthor{arjovsky2015unitary} to allow for $n^2$ parameters may result in performance similar to our approach. However, how to go about such a generalisation is unclear, as a naive approach would simply use a composition of $n^2$ operators, and this would likely become computationally intractable.

\subsection{Method of generating $U$}
\label{section:method}
As described, we used three methods to generate the true $U$.
One of these produces $U$ in the subspace available to the composition parametrization (Equation~\ref{eqn:arjovsky}), so we were curious to see if this parametrization performed better on experiments using that method.
We were also concerned that generating $U$ using the Lie algebra parametrization might make the task too `easy' for our approach, as its random initialization could lie close to the true solution. 

\begin{figure}
\centering
\includegraphics[scale=0.55]{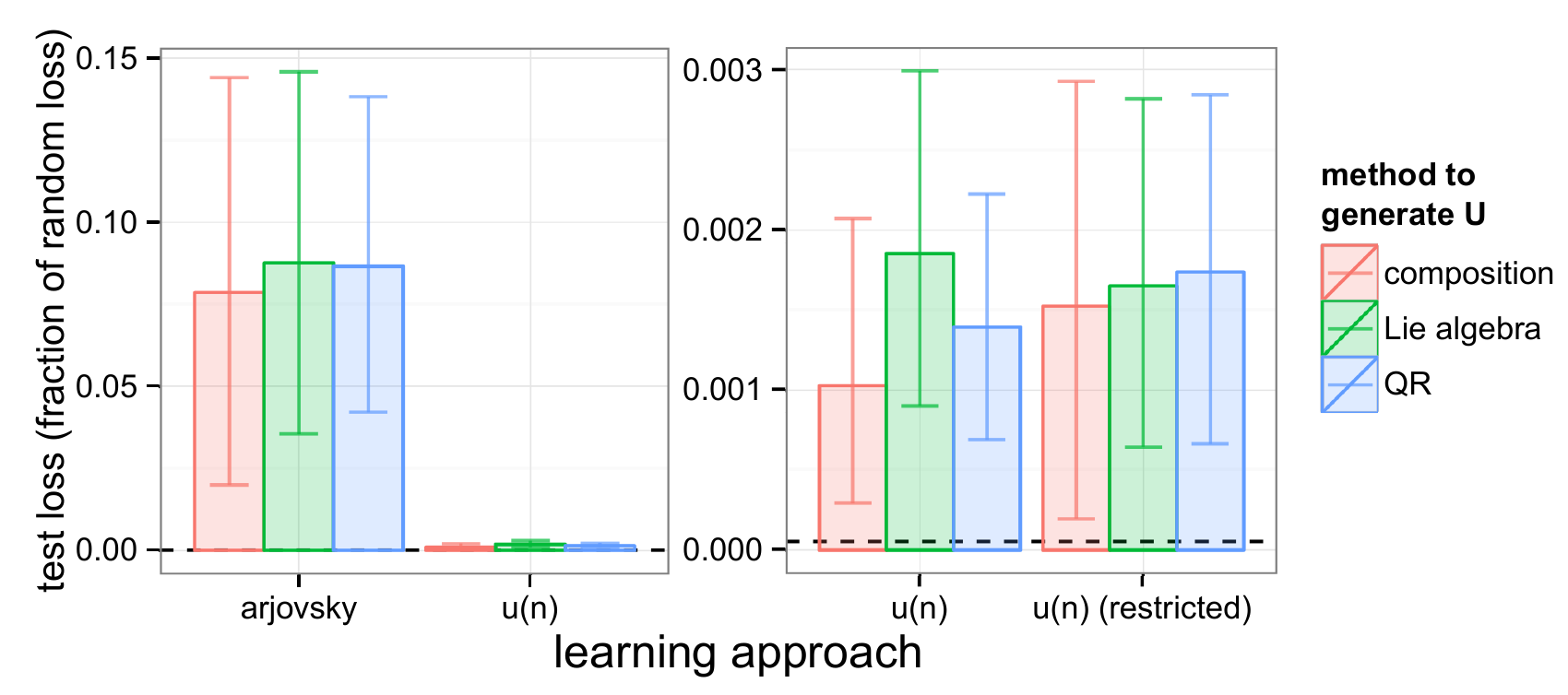}
\caption{We ask whether the method used to generate $U$ influences performance for different approaches to \emph{learning} $U$.
Error bars are bootstrap estimates of 95\% confidence intervals. To compare across different $n$'s, we normalise each loss by the loss of \texttt{rand} for that $n$, and reporrt fractions.
The dotted line is the \texttt{true} loss, similarly normalised.
the choice of method to generate $U$ does not appear to affect test-set loss for the different approaches. \emph{Right}: Finer resolution on the $\mathfrak{u}(n)$ result in left panel. We also include the case where we restrict to $7n$ learnable parameters.}
\label{fig:boxplots}
\end{figure}

Figure~\ref{fig:boxplots} shows box-plots of the distribution of test losses from these approaches for the three methods, comparing our approach ($\mathfrak{u}(n)$) with that of \citeauthor{arjovsky2015unitary}~\shortcite{arjovsky2015unitary}, denoted \texttt{arjovsky}.
To combine results from experiments using different values of $n$, we first scaled test-set losses by the performance of \texttt{rand} (the random unitary matrix), so the y-axis ranges from 0 (perfect) to 1 (random performance).
The dotted line denotes the average (over methods) of the test-set loss for \texttt{true}, similarly scaled.
The right panel in Figure~\ref{fig:boxplots} shows a zoomed-in version of the $\mathfrak{u}(n)$ result where the comparison with \texttt{true} is more meaningful, and a comparison with the case where we have restricted to $7n$ learnable parameters (see earlier).

We do not observe a difference (within error) between the methods, which is consistent between $\mathfrak{u}(n)$ and \texttt{arjovsky}.
Our concern that using the Lie algebra to generate $U$ would make the task `too easy' for $\mathfrak{u}(n)$ was seemingly unfounded.

\section{Unitary Recurrent Neural Network for Long Memory Tasks}
To demonstrate that our approach is practical for use in deep learning, we incorporate it into a recurrent neural network to solve standard long-memory tasks. Specifically, we define a general unitary RNN with recurrence relation
\begin{equation}
\mathbf{h}_t = f\left( \beta U \mathbf{h}_{t-1} + V \mathbf{x}_t + \mathbf{b} \right)
\end{equation}
where $f$ is a nonlinearity, $\beta$ is a free scaling factor, $U$ is our unitary matrix parametrised as in equation~\ref{eqn:parametrization}, $\mathbf{h}_t$ is the hidden state of the RNN and $\mathbf{x}_t$ is the input data at `time-point' $t$. We refer to this as a `general unitary RNN' (guRNN), to distinguish it from the restricted uRNN of~\citeauthor{arjovsky2015unitary}~\shortcite{arjovsky2015unitary}.

\begin{figure}
\includegraphics[scale=0.85]{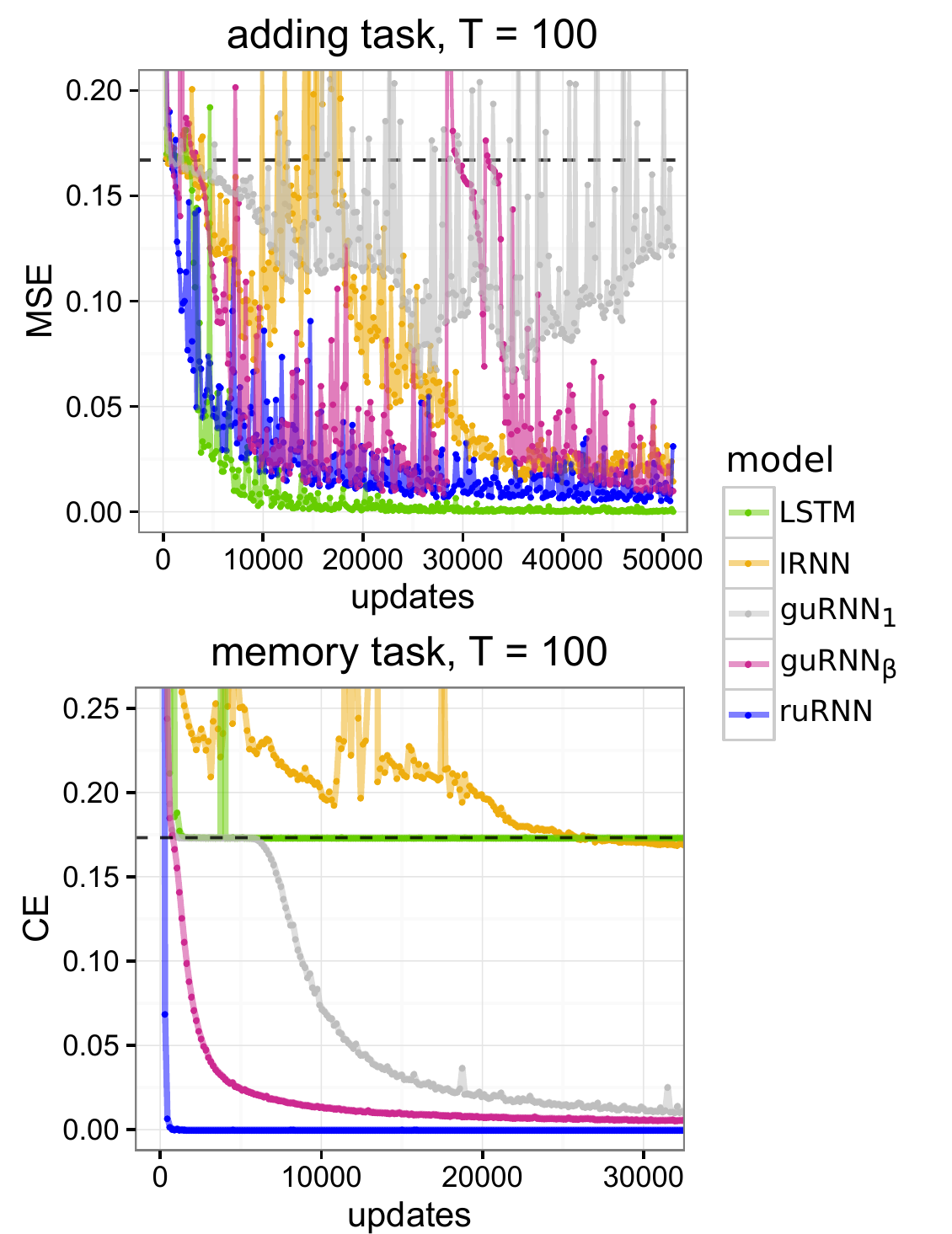}
\caption{We compare different RNN models on two standard long-memory learning tasks, described in~\cite{hochreiter1997long}. The state size for all models was $n=30$, except for the ruRNN~\cite{arjovsky2015unitary}, which had $n=512$ and $n=128$ for the adding and memory tasks respectively; the optimal hyperparameters reported in their work. For our model (guRNN), we use $f=\mathrm{relu}$ 
 and $f=\mathrm{tanh}$ for the nonlinearities in the adding and memory tasks. We compare guRNN with (guRNN$_\beta$) or without (guRNN$_1$) a scaling factor $\beta$ in front of $U$ to compensate for the tendency of the nonlinearity to shrink gradients. We used $\beta = 1.4$ and $\beta = 1.05$. That $\mathrm{relu}$ requires a larger $\beta$ is expected, as this nonlinearity discards more gradient information. Gradient clipping to $[-1, 1]$ was used for the LSTM and IRNN~\cite{le2015simple}. Dotted lines denote random baselines. The learning rate was set to $\alpha=10^{-3}$ for all models except IRNN, which used $\alpha=10^{-4}$. We used RMSProp~\cite{Tieleman2012} with decay 0.9 and no momentum. The batch size was 20.}
\label{fig:RNN}
\end{figure}

We use the guRNN on two tasks: the `adding problem' and the `memory problem', first described in~\cite{hochreiter1997long}. For the sake of brevity we refer to~\cite{arjovsky2015unitary} for specific experimental details, as we use an identical experimental setup (reproduced in TensorFlow; see above github link for code). We compare our model (guRNN) with the restricted uRNN (ruRNN) parametrised as in equation~\ref{eqn:arjovsky}, a LSTM~\cite{hochreiter1997long}, and the IRNN of~\citeauthor{le2015simple}~\shortcite{le2015simple}. Figure~\ref{fig:RNN} shows the results for each task where the sequence length or the memory duration is $T=100$.

While our model guarantees unitarity of $U$, this is \emph{not} sufficient to prevent gradients from vanishing. Consider the norm of the gradient of the cost $C$ with respect to the data at time $\tau$, and use submultiplicativity of the norm to write;

\begin{equation*}
\left\| \frac{\partial C}{\partial \mathbf{x}_\tau} \right\| \leq \left\| \frac{\partial C}{\partial \mathbf{x}_T} \right\| \left( \prod_{t = \tau}^{T-1} \| f' \left( U \mathbf{h}_t + V \mathbf{x}_t + \mathbf{b}\right) \| \| U \|\right) \left\|\frac{\partial \mathbf{h}_\tau}{\mathbf{x}_\tau} \right\|
\end{equation*}
where $f'$ is a diagonal matrix giving the derivatives of the nonlinearity. Using a unitary matrix fixes $\| U \| = 1$, but beyond further restrictions (on $V$ and $\mathbf{b}$) does nothing to control the norm of $f'$, which is at \emph{most} 1 for common nonlinearities. Designing a nonlinearity to better preserve gradient norms is beyond the scope of this work, so we simply scaled $U$ by a constant multiplicative factor $\beta$ to counteract the tendency of the nonlinearity to shrink gradients. In Figure~\ref{fig:RNN} we denote this setup by guRNN$_\beta$. Confirming our intuition, this simple modification greatly improves performance on both tasks. 

Perhaps owing to our efficient gradient calculation (appendix A) and simpler recurrence relation, our model runs faster than that of~\cite{arjovsky2015unitary} (in our implementation), by a factor of 4.8 and 2.6 in the adding and memory tasks shown in Figure~\ref{fig:RNN} respectively. This amounts to the guRNN processing 61.2 and 37.0 examples per second in the two tasks, on a GeForce GTX 1080 GPU.

\section{Discussion}
\label{section:discussion}
Drawing from the rich theory of Lie groups and Lie algebras, we have described a parametrization of unitary matrices appropriate for use in deep learning.
This parametrization exploits the Lie group-Lie algebra correspondence through the exponential map to represent unitary matrices in terms of real coefficients relative to a given basis of the Lie algebra $\mathfrak{u}(n)$.
As this map from $\mathfrak{u}(n)$ to $U(n)$ is surjective, the parametrization can describe any unitary matrix.

We have demonstrated that unitary matrices can be learned with high accuracy using simple gradient descent, and that this approach outperforms a recently-proposed parametrization (from~\citeauthor{arjovsky2015unitary}~\shortcite{arjovsky2015unitary}) and significantly outperforms the approach of `re-unitarizing' after gradient updates. This experimental design is quite simple, designed to probe a core problem, before considering the broader setting of RNNs.

Our experiments with general unitary RNNs using this parametrization showed that this approach is practical for deep learning. With a fraction of the parameters, our model outperforms LSTMs on the standard `memory problem' and attains comparable (although inferior) performance on the adding problem~\cite{hochreiter1997long}. Further work is required to understand the difference in performance between our approach and the ruRNN of~\cite{arjovsky2015unitary} - perhaps the $7n$-dimensional subspace captured by their parametrization is serendipitously \emph{beneficial} for these RNN tasks - although we note that the results presented here are not the fruit of exhaustive hyperparameter exploration. Of particular interest is the impressive performance of both uRNNs on the memory task, where the LSTM and IRNN appear to fail to learn.

While our RNN experiments have demonstrated the utility of using a unitary operator for these tasks, we believe that the role of the nonlinearity in the vanishing and exploding gradient problem must not be discounted. We have shown that a simple scaling factor can help reduce the vanishing gradient problem induced by the choice of nonlinearity. More analysis considering the combination of nonlinearity and transition operator must be performed to better tackle this problem.

The success of our parametrization for unitary operator learning suggests that the approach of performing gradient updates in the Lie algebra is particularly effective. As Lie groups describe many naturally-occuring symmetries, the Lie group-Lie algebra correspondence could be rich for further exploitation to enhance performance in tasks beyond our initial motivation of recurrent neural networks.

\section{Appendix A: Derivation of derivative of the matrix exponential}
This derivation draws elements from~\citeauthor{kalbfleisch1985analysis}~\shortcite{kalbfleisch1985analysis} and~\citeauthor{jennrich1976fitting}~\shortcite{jennrich1976fitting}.

We have $U = \exp(L)$, and seek $dU$. For what follows, we simply require that $L$ be normal, so the results are more general than the unitary case. In this case, $L$ is skew-Hermitian, which is normal and therefore diagonalisable by unitary matrices. Thus, there exist $W \in U(n)$ and $D = \mathrm{diag}(d_1, \dots, d_n)$ such that $L = W D W^{\dagger}$, and therefore
\begin{equation}
U = W \tilde{D} W^{\dagger}
\label{eqn:-1}
\end{equation}
where $\tilde{D} = \mathrm{diag}(e^{d_1}, \dots, e^{d_n})$.

We assume we can calculate: $dL$, $W$, and $D$ and seek an expression for $dU$.

Then using~\ref{eqn:-1}:
\begin{equation}
\begin{split}
dU &= d(W \tilde{D} W^{\dagger}) \\
&= dW \tilde{D} W^{\dagger} + W d\tilde{D} W^{\dagger} + W \tilde{D} dW^{\dagger}
\end{split}
\end{equation}

Pre-multiplying with $W^{\dagger}$ and post-multiplying with $W$:
\begin{equation}
W^{\dagger} dU W = W^{\dagger} dW \tilde{D} + d\tilde{D} + \tilde{D} dW^{\dagger} W
\label{eqn:0}
\end{equation}

The last term can be simplified by differentiating both sides of $W^{\dagger} W = \mathbb{I}$ (this follows from unitarity of $W$);
\begin{equation}
W^{\dagger} W + W^{\dagger} dW = 0 \Rightarrow dW^{\dagger} W = -W^{\dagger} dW
\label{eqn:1}
\end{equation}
and substituting back into~\ref{eqn:0} to get:
\begin{equation}
W^{\dagger} dU W = W^{\dagger} dW \tilde{D} - \tilde{D} W^{\dagger} dW + d\tilde{D} 
\end{equation}

We can then say that $dU = W V W^{\dagger}$ where
\begin{equation}
V = W^{\dagger} dW \tilde{D} - \tilde{D} W^{\dagger} dW + d\tilde{D}
\end{equation}

Similarly, $dL = W A W^{\dagger}$ where (replacing $\tilde{D}$ with $D$)
\begin{equation}
A = W^{\dagger} dW D - D W^{\dagger} dW + dD
\label{eqn:2}
\end{equation}
and also $A = W^{\dagger} dL W$.

\subsection{Calculating $V$}
We use the convention that repeated indices denote summation over that index, unless otherwise stated.

Looking at the components of $V$;
\begin{equation}
V_{ij} = (W^{\dagger} dW \tilde{D})_{ij} - (\tilde{D} W^{\dagger} dW)_{ij} + d\tilde{D}_{ij}
\end{equation}

\subsubsection{Diagonal case ($i = j$): (no summation over $i$)}
\begin{equation}
V_{ii} = W^{\dagger}_{ia} dW_{ab} \tilde{D}_{bi} - \tilde{D}_{ia} W^{\dagger}_{ab} dW_{bi} + d\tilde{D}_{ii}
\end{equation}
Since $\tilde{D}_{bi} = \delta_{bi} \tilde{d}_i$, the first two terms cancel:
\begin{equation}
\begin{split}
V_{ii} =& W^{\dagger}_{ia} dW_{ab} \delta_{bi} \tilde{d}_i - \delta_{ai} \tilde{d}_i W^{\dagger}_{ab} dW_{bi} + d\tilde{D}_{ii}\\
=& W^{\dagger}_{ia} dW_{ai} \tilde{d}_i - \tilde{d}_i W^{\dagger}_{ib} dW_{bi} + d\tilde{D}_{ii}\\
=& d\tilde{D}_{ii}
\label{eqn:3}
\end{split}
\end{equation}
Using~\ref{eqn:2} we get $A_{ii} = dD_{ii} = (W^{\dagger} dL W)_{ii}$

Recall that the diagonal elements of $\tilde{D}$ are the exponentiated versions of the diagonal elements of $D$, so $\tilde{D}_{ii} = e^{d_i}$. Then
\begin{equation}
d\tilde{D}_{ii} = d(d_i) e^{d_i} = dD_{ii} \tilde{D}_{ii}
\end{equation}

Inserting that into Equation~\ref{eqn:3}:
\begin{equation}
V_{ii} = dD_{ii} \tilde{D}_{ii} = (W^{\dagger} dL W)_{ii} \tilde{D}_{ii} = (W^{\dagger} dL W)_{ii} e^{d_i}
\end{equation}
This produces Equation~\ref{eqn:vii} in the main paper.

\subsubsection{Off-diagonal case ($i \neq j$): (no summation over $i, j$)}
In this case, the purely diagonal part vanishes. We get:
\begin{equation}
\begin{split}
V_{ij} =& W^{\dagger}_{ia} dW_{ab} \delta_{bj} \tilde{d}_j - \delta_{ai} \tilde{d}_i W^{\dagger}_{ab} dW_{bj}\\
=& W^{\dagger}_{ia} dW_{aj} \tilde{d}_j - W^{\dagger}_{ib} dW_{bj} \tilde{d}_i \\
&= (W^{\dagger} dW)_{ij} (\tilde{d}_j - \tilde{d}_i)
\end{split}
\label{eqn:previj}
\end{equation}
Similarly,
\begin{equation}
A_{ij} = (W^{\dagger} dW)_{ij} (d_j - d_i)
\end{equation}
Remembering that this is all component-wise multiplication (no summation over $i$ and $j$), we can rearrange expressions to get:
\begin{equation}
(W^{\dagger} dW)_{ij} = \frac{A_{ij}}{d_j - d_i} = \frac{(W^{\dagger} dL W)_{ij}}{d_j - d_i}
\end{equation}
Combining this with~\ref{eqn:previj} and remembering $\tilde{d}_a = e^{d_a}$, we have, for $i \neq j$:
\begin{equation}
V_{ij} = (W^{\dagger} dL W)_{ij} \left(\frac{e^{d_i} - e^{d_j}}{d_i - d_j}\right)
\end{equation}
This is Equation~\ref{eqn:vij} in the main paper.

\subsection{Efficiently calculating $W^{\dagger} dL W$}
This section is specific to our work, as it relies on the choice of basis for $\mathfrak{u}(n)$. 

In our case, $dL$ is simple. $L$ is a linear combination of the parameters $\lambda_i$;
\begin{equation}
L = \sum_i^{n^2} \lambda_i T_i
\end{equation}

Where $T_i$ are the basis matrices of $\mathfrak{u}(n)$.

Then
\begin{equation}
dL_a = \frac{\partial L}{\partial \lambda_a} = T_a
\end{equation}

We need $W^{\dagger} T_a W$ for all $a$. Since the $T_a$s are sparse, this is cheaper than performing $n^2$ full matrix multiplications, as we demonstrate now.

In components;
\begin{equation}
(W^{\dagger} T_a W)_{i j} = W^{\dagger}_{i k} {T_a}_{k l} W_{l j}
\end{equation}

Cases:
\subsubsection{$T_a$ diagonal, purely imaginary}
$T_a$ is zero except for a $i$ in the $a$-th position on the diagonal.
\begin{equation}
\begin{split}
(W^{\dagger} T_a W)_{i j} &= i W^{\dagger}_{i a} W_{a j} = i W^*_{a i} W_{a j}\\
\Rightarrow W^{\dagger} T_a W &= i \cdot \mathrm{outer}(\mathbf{w}^*_a, \mathbf{w}_a)
\end{split}
\end{equation}
where $\mathbf{w}_a$ is the $a$-th \emph{row} of $W$.

\subsubsection{$T_a$ symmetric, purely imaginary}
$T_{rs}$ is zero except for $i$ in position $(r, s)$ and $(s, r)$.
\begin{equation}
\begin{split}
(W^{\dagger} T_{rs} W)_{i j} &= i W^{\dagger}_{i k} (\delta_{ks, lr} + \delta_{kr, ls}) W_{l j} \\
&= i (W^{\dagger}_{is} W_{rj} + W^{\dagger}_{ir} W_{s j})  = i ( W^*_{si} W_{rj} + W^*_{ri} W_{sj}) \\
\Rightarrow W^{\dagger} T_{rs} W &= i \cdot (\mathrm{outer}(\mathbf{w}^*_s, \mathbf{w}_r) + \mathrm{outer}(\mathbf{w}^*_r, \mathbf{w}_s))
\end{split}
\end{equation}

\subsubsection{$T_a$ antisymmetric, purely real}
$T_{rs}$ is zero except for $1$ in position $(r, s)$ and $-1$ in position $(s, r)$.
\begin{equation}
\begin{split}
(W^{\dagger} T_{rs} W)_{i j} &= W^{\dagger}_{i k} (\delta_{kr, sl} - \delta_{ks, rl}) W_{l j} \\
&= W^{\dagger}_{ir} W_{sj} - W^{\dagger}_{is} W_{r j}  =  W^*_{ri} W_{sj} - W^*_{si} W_{rj} \\
\Rightarrow W^{\dagger} T_{rs} W &= \mathrm{outer}(\mathbf{w}^*_r, \mathbf{w}_s) - \mathrm{outer}(\mathbf{w}^*_s, \mathbf{w}_r)
\end{split}
\end{equation}

These reproduce the expressions in the main paper. The outer product of two $n$-dimensional vectors is an $O(n^2)$ operation, and so this provides a (up to) factor $n$ speed-up on matrix multiplication.

\section{Appendix B: Changing the basis of $\mathfrak{u}(n)$}
\label{section:basis}
\begin{figure}
\includegraphics[scale=0.5]{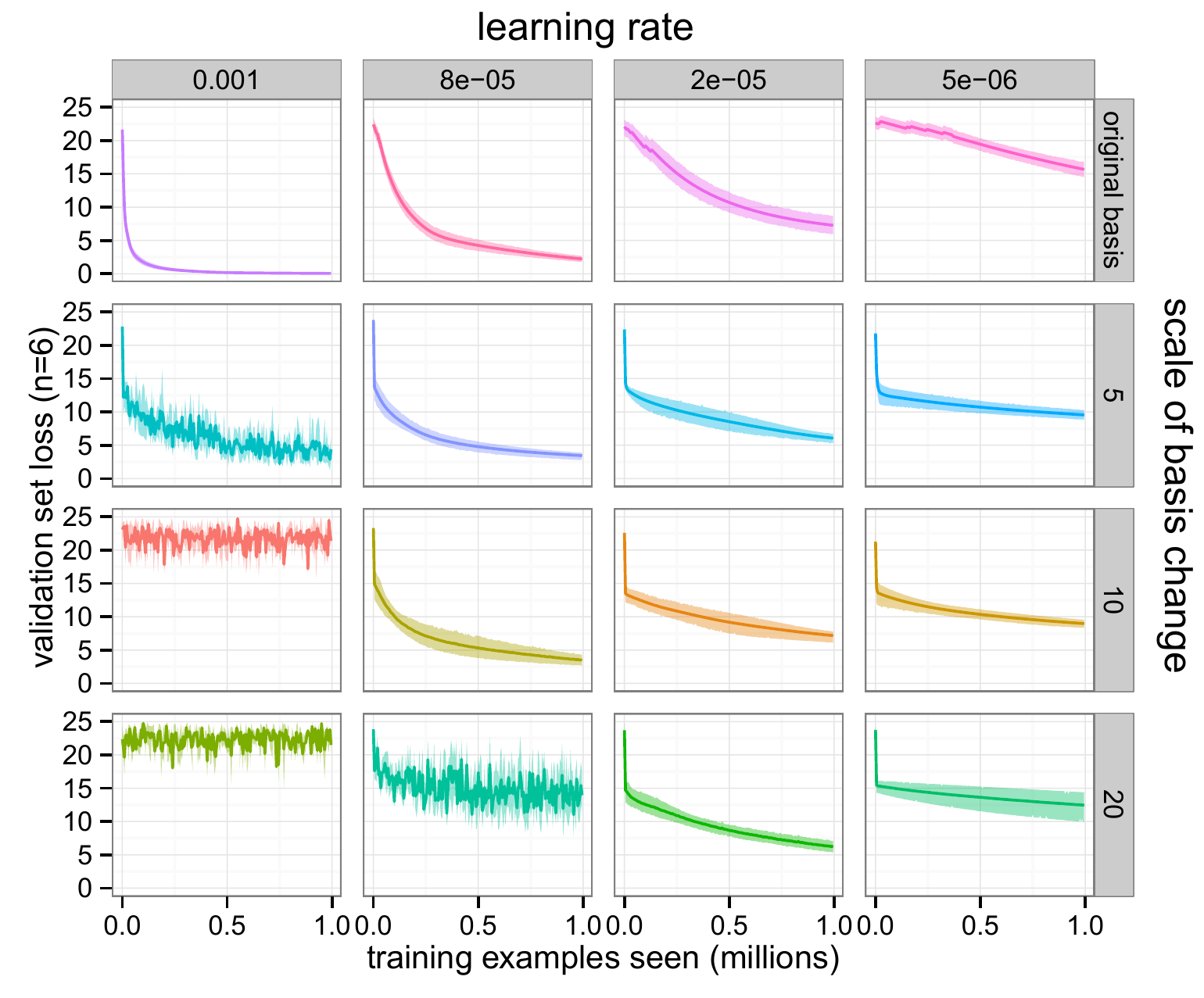}
\caption{
We consider the effects on learning of changing the basis (rows) and changing the learning rate (columns). For this experiment, $n=6$. The first row uses the original basis. Other rows use change of basis matrices sampled from $\mathcal{U}(-c, c)$ where $c = \{5, 10, 20\}$. The learning rates decrease from the `default' value of 0.001 used in the other experiments. Subsequent values are given by $\frac{0.001}{c^2}$ for the above values of $c$, in an attempt to rescale by the expected absolute value of components of the change of basis matrix. If the change of scale were solely responsible for the change in learning behavior, we would therefore expect the graphs on the diagonal to look the same.}
\label{fig:basis}
\end{figure}

The Lie group parametrization assumes a fixed basis of $\mathfrak{u}(n)$. 
Our intuition is that this makes some regions of $U(n)$ more `accessible' to the optimization procedure, elements whose coefficients are small given this basis. 
Learning a matrix $U$ which came from elsewhere in $U(n)$ may therefore be more challenging.
We emulated this `change of basis` without needing to explicitly construct a new basis by generating a change of basis matrix, $M$.
That is, if $V_j$ is the $j$-th element of the new basis, it is given by
\begin{equation}
V_j = \sum_k M_{jk} T_k
\end{equation}
If $\{\tilde{\lambda}\}_a$ are the coefficients of $L$ relative to the basis $V$, the coefficients relative to the old basis $T$ are given by:
\begin{equation}
\lambda_b = \sum_k \tilde{\lambda}_k M_{kb} = \tilde{\lambda}^T \cdot M
\end{equation}
A change of basis matrix must be full-rank.
We generate one by sampling a square, $n^2 \times n^2$ matrix from a continuous uniform distribution $\mathcal{U}(-c, c)$ ($c$ is a constant we vary in experiments, see Figure~\ref{fig:basis}).
This is very unlikely to be singular.
We choose the $c$ range of the distribution such that $M$ will have `large' values relative to the true matrix $U$, whose parameters $\lambda$ (relative to $T$) are drawn from $\mathcal{N}(0, 0.01)$.

Preliminary experiments suggested that the learning rate must be adjusted to compensate for the change of scale - evidence for this is visible in the first column of Figure~\ref{fig:basis}, where changing the basis without changing the learning rate results in an unstable validation set trace.
Poor performance resulting from an inappropriate learning rate is not our focus here, so we performed experiments for different values of the learning rate.
Figure~\ref{fig:basis} shows a grid of validation set losses as we vary the learning rate (columns) and the value of $c$ (rows).

Our intuition is that if the performance under the change of basis is \emph{purely} driven by the difference in scale, using an appropriately-scaled learning rate should negate its affect.
Each parameter $\lambda_j$ is scaled by a variable uniformly distributed between $(-c, c)$.
The expectation value of the absolute value of this quantity is $c^2/2$, so we consider learning rates normalised by this factor. 

As seen in Figure~\ref{fig:basis}, the graphs on the diagonal are \emph{not} identical, suggesting that merely scaling the learning rate does not account for the change of learning behavior given a new basis - at least in expectation.
Nonetheless, it is reassuring to observe that for all choices of $c$ explored, there exists a learning rate which facilitates learning, even if it markedly slower than the `ideal' case.
While having a `misspecified' basis does appear to negatively impact learning, it can be largely overcome with choice of learning rate.

\bibliographystyle{aaai}
\bibliography{hyland_aaai_crc}

\end{document}